# MONITORING THE IMPACTS OF A TAILINGS DAM FAILURE USING SATELLITE IMAGES

Jaime Moraga, PhD[1]
Gurbet Gurkan, MSc[2]
Prof. Dr. H. Sebnem Duzgun[3]

## ABSTRACT

Monitoring dam failures using satellite images provides first responders with efficient management of early interventions.  It is also equally important to monitor spatial and temporal changes in the inundation area to track the post-disaster recovery.  On January 25th, 2019, the tailings dam of the Córrego do Feijão iron ore mine, located in Brumadinho, Brazil, collapsed. This disaster caused more than 230 fatalities and 30 missing people leading to damage on the order of multiple billions of dollars.

This study uses Sentinel-2 satellite images to map the inundation area and assess and delineate the land use and land cover impacted by the dam failure. The images correspond to data captures from January 22nd (3 days before), and February 02 (7 days after the collapse).

Satellite images of the region were classified for before and aftermath of the disaster implementing a machine learning algorithm. In order to have sufficient land cover types to validate the quality and accuracy of the algorithm, 7 classes were defined: mine (mining pit, stockpiles, tailings dam), forest, build up (buildings, houses, roads), river, agricultural (plots of land that are either being cultivated or agricultural bare land waiting to be cultivated), clear water (lakes and reservoirs), and grassland (non-forest natural vegetation and yards/parks).

The developed classification algorithm yielded a high accuracy (99%) for the image before the collapse. This paper determines land cover impact using two different models, 1) by using the trained network in the "after" image (an approach that can be used as a low cost and first assessment of the impact), and 2) by creating a second network, trained in a subset of points of the "after" image, and then comparing the land cover results of the two trained networks (requires relabeling part of the new data and can delay results, but provides better outcomes).  In the first model, applying the trained network to the "after" image, the accuracy is still high (86%), but lower than using the second model (98%).

This strategy can be applied at a low cost for monitoring and assessment by using openly available satellite information and, in case of dam collapse or with a larger budget, higher resolution and faster data can be obtained by fly-overs on the area of concern.

[1] Colorado School of Mines, Golden, Colorado, jmoraga@mines.edu
[2] Colorado School of Mines, Golden, Colorado, ggurkan@mines.edu
[3] Colorado School of Mines, Fred Banfield Distinguished Endowed Chair in Mining Engineering, Golden, Colorado, duzgun@mines.edu

# INTRODUCTION

During 2019, there have been three tailings dam collapses in Brazil: Córrego de Feijão mine, an iron mine in Brumadinho, Região Metropolitana de Belo Horizonte , in the state of Minas Gerais Jan 25, 2019); Machadinho d'Oeste, a tin mine in Oriente Novo, in the state of Rondônia (March 29, 2019), and; a gold mine in Nossa Senhora do Livramento , in the state of Mato Grosso (Oct 1, 2019).

The most devastating one occurred in the Córrego do Feijão iron ore mine, located in Brumadinho on January 25th, 2019. This disaster caused more than 230 fatalities and 30 missing people, leading to damage in the order of multiple billions of dollars (BBC News, 2019). After such a disaster it is important to know the impacted area to make an effective assessment and take immediate measures for relief and recovery. For this purpose, remote sensing can be the most effective and the fastest practice in disaster management (Bello and Aina, 2014), which can assist in damage assessment and aftermath monitoring, providing a quantitative base for relief operations (Westen, 2000). This paper applies a remote sensing methodology based exclusively on openly available multispectral satellite imaging, but companies could use a mix of free-access and privately generated data to increase accuracy and speed of monitoring.

Application of machine learning algorithms on remote sensing has been used to improve the results of classification. There are many studies focused on mapping the affected area after a disaster by using combination of satellite imagery with machine learning algorithms. For instance, Syifa et al. (2019) studied the Brumadinho dam collapse in Brazil to map and calculate the dimensions of the flood caused by the collapse using remote sensing. They made a pixel-based classification for the pre- and post-flood images from Landsat-8 and Sentinel-2 applying two different artificial intelligence techniques: artificial neural network (ANN) and support vector machines (SVM).

These classifiers were able to determine and calculate the flood area successfully. Luo et al. (2019) analyzed the Bento Rodrigues dam disaster in Brazil by using Landsat ETM+ and OLI images of the disaster area. They tested the performance of SVM and decision tree (DT) classifiers for mapping the changes in land cover caused by the disaster and, instead of just using machine learning algorithms directly on the images, evaluated the effect of adding normalized difference vegetation index (NDVI) as another layer of input data on top of the multi-spectral image from the satellite images. They concluded that NDVI can improve the vegetated land cover types image processing, as opposed to using NDVI or multi-spectral images independently. Besides, SVM gave higher accuracy than DT in classification of land use and land cover map.

Junior et al. (2018) combined Unmixing Espectral Linear Model (UELM), artificial neural network, Enhanced Vegetation Index (EVI) and Normalized Difference Vegetation Index (NDVI) to determine the change in vegetation cover in the 2015 Fundão Dam failure, also in Brazil, by using Landsat-8 images. UELM was used to separate soil, shade, and vegetation classes. After that ANN was applied for classification followed by EVI and NDVI. With this classification, they found out a loss of vegetation of 13.02% occurred as a result of the disaster.

In this study, convolutional neural networks (CNN), one of the most useful machine learning algorithms in classification of images, are used to create land cover maps in order to detect the affected area after the Brumadinho tailings dam disaster.

The approach combines the use of a CNN tailored for analysis of multispectral images for Land Use Land Cover, LULC by Moraga, et al, (2019), a data gathering and data augmentation approach, and application-specific analysis that takes advantage of the characteristics of the problem to obtain highly accurate maps of the affected areas that include type of area affected.

## METHODOLOGY

This paper uses a CNN tailored for analysis and classification of LULC, the Jigsaw CNN (Moraga, et al, 2019), to classify each point in a satellite image of the impacted area. Based on the classifications made for the before and after images, a change analysis is performed to identify affected area and changes in the LULC.

Figure 1 describes the architecture of the Jigsaw CNN. It consists of more than 20 layers organized in 3 blocks. The source data is preprocessed and turned into 17x17 tiles that will become the input to be processed in two parallel structures (A&B), and the results merged and classified in block C.

Block A applies convolutions to the image, looking for features and patterns in kernels of sizes 3x3, 5x5 and 7x7, also applying max pooling and a regular convolution to later turn all the results into a single vector for further processing. This results in a classification that identifies patterns that are area specific, but may lose definition or information that is present in each pixel.

Block B, on the other hand, uses two layers of dense networks to extract information from each multi-spectral pixel in the image.

The concatenated outputs of A & B are processed together in Block C, where an activation function turns them into a class that should match the LULC labels.

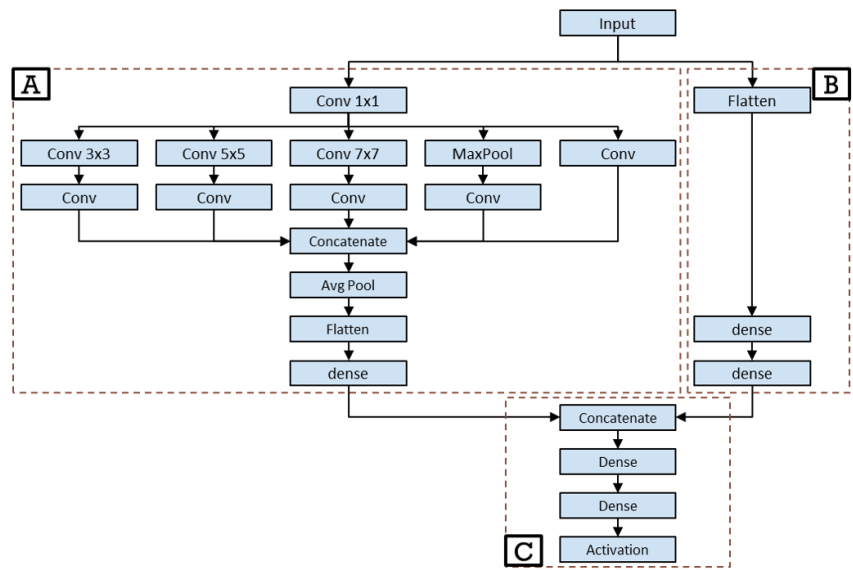

**Figure 1. Jigsaw architecture, based on Inception**

The approach in Figure 2 shows the process of acquisition, annotation, classification and evaluation of results, streamlining the effort required to evaluate the area affected and what type of impact the dam failure had in the region. Figure 2 illustrates the process used to train the network.

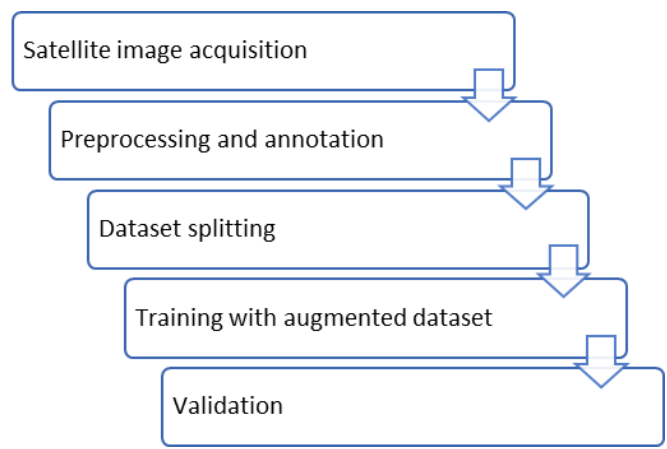

**Figure 2. Jigsaw Network Training Process**

For the input data, pre and post disaster multispectral images were acquired from Sentinel-2 (Figure 3). Sentinel-2 gives global coverage every five days and is equipped with multispectral imager (MSI) with 13 bands (Drusch et al., 2012). This limits the practical application of this approach in real life, because it may take days before an image is captured and published by ESO on a specific affected area.

The spatial resolution of the MSI varies between 10m, 20m, and 60m for different bands (European Space Agency, 2019), so the image was resampled to 10m by 10m pixels by slicing the images with larger resolution into smaller pixels (no interpolation was applied).

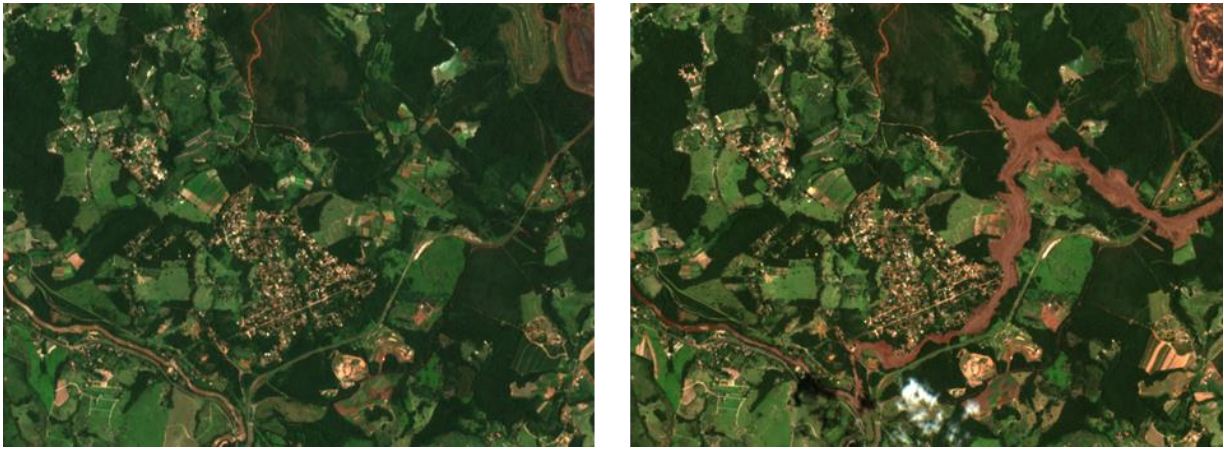

**Figure 3. Visible spectra of the images used. Left is before the collapse, and right is after the collapse of the tailings dam**

To annotate the image, small representative were selected for seven relevant land use classes (Table 1).

**Table 1. Classes used for annotation, color used on maps, and class description**

| Class | Name | Color | Description |
|---|---|---|---|
| 1 | Mine & tailings | Red | Mining pit, stockpiles, tailings dam |
| 2 | Forest | Green | Large masses of trees either natural or man-made |
| 3 | Build up | Yellow | buildings, houses, roads |
| 4 | River | Blue | Rivers |
| 5 | Clear water | Cyan | Lakes and reservoirs |
| 6 | Agricultural | Purple | Plots of land that are either being cultivated or agricultural bare land waiting to be cultivated |
| 7 | Grassland | White | Non-forest natural vegetation and yards/parks |

The input for the network consists of small crops of the images – the pieces of the "jigsaw" (Moraga et al. 2019) – and to create those the image was split in 17 by 17 pixel tiles or blocks (each labeled by the class of their center pixel as shown in Figure 4), and augment those blocks by rotation and mirroring to create a robust dataset to train the image classification network. Each pixel in the image can then be recognized independently by using not only its information but also the region surrounding it. This increases the robustness of the solution and allows more granularity in the classification by using region-specific characteristics (for example, if a green region is surrounded by buildings, it will not be classified as a forest).

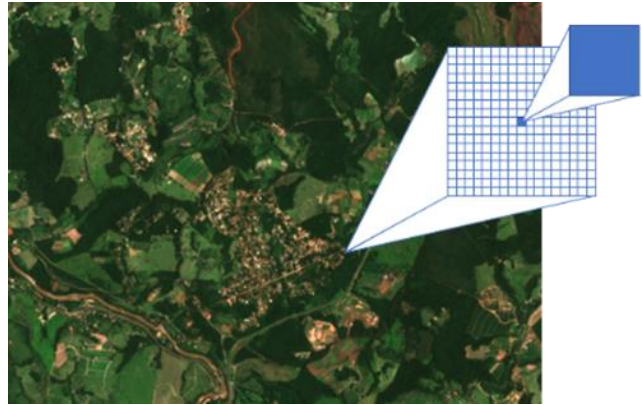

**Figure 4. The image is split by creating 17 by 17 blocks centered on each pixel. Blocks will be classified by the class of their middle pixel.**

Finally, a consolidated map of the region affected by the tailings was built: by using the trained network on both images and comparing the results. This analysis is not limited to the labeled areas of the images, but comprises the totality of each image, to allow for later visual confirmation of the areas affected.

Because the tailings recognition has a very high degree of accuracy, this class can be used as a mask to identify the types of land use that were affected by the tailings. The stages of the process are given in Figure 5.

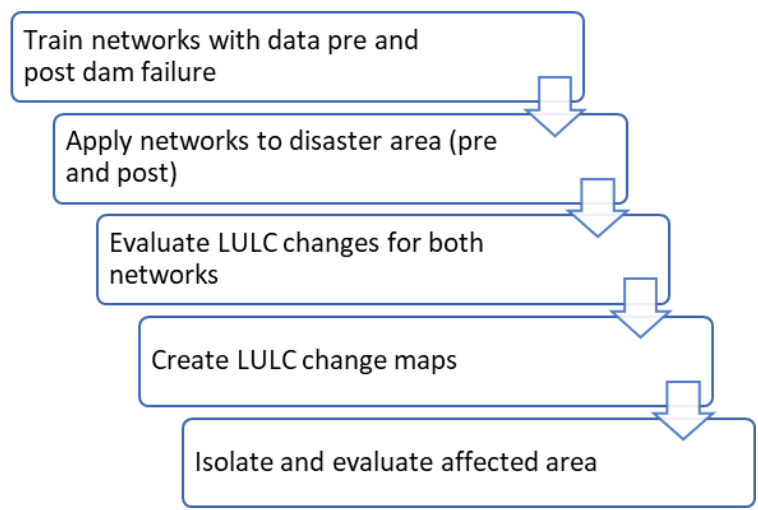

**Figure 5. The process to create the change maps is streamlined through automation**

# CHANGE MAP AND RESULTS

By using Jigsaw on 1,200 samples of each class of pixel (8,400 tiles), a network was trained with a random sample of 50% of all samples (4,200 tiles). Using this trained network, two images were generated with classification of land-use and land-cover: one for before the event (the "before" image) and one for after the event (the "before" image). The results of the classification have a high degree of accuracy, with the networks trained with the image pre-collapse having 99%, and the one trained with the image after the collapse achieving a 98% accuracy (measured against the test set corresponding to the 4,200 tiles that were not used for testing).

The images are depicted on Figure 6. The classification in both cases is quite robust, and the effect of the dam collapse can be seen on the expansion of the red (Mine and Tailings) area.

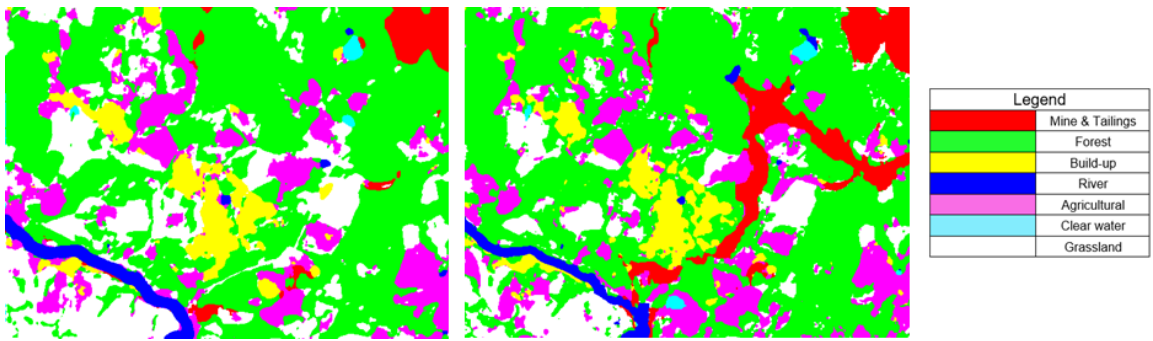
**Figure 6. Results of classification for "before" and "after" images**

The accuracy scores for each network and class are included as confusion matrices in Table 2 and Table 3. The confusion matrices use the results and expected classification of the test data as a source. The matrix shows which percentage of each sample was classified as each class. And accuracy per class can be seen in the diagonal (sample was classified as the expected class).
It's important to note that:
   a) Mine & Tailings accuracy is high for both cases. For the after images, it drops to 98.4%, which is expected due to the more complex spread of the tailings
   b) The largest uncertainty comes from classification of green areas: agricultural, forest and grasslands. This is expected, as there is a semantic component related to the type of vegetation
   c) Given that forest is the largest area, it is expected that this uncertainty will be overrepresented
   d) Built-up, and river are classified accurately on both networks

**Table 2. Confusion matrix after classification of "before" image**

| | | Actual Value | | | | | | |
|---|---|---|---|---|---|---|---|---|
| | | River | Forest | Clear water | Mine + Tailings | Agricultural | Built-up | Grassland |
| **Predicted** | River | **100%** | 0.0% | 0.0% | 0.0% | 0.0% | 0.0% | 0.0% |
| | Forest | 0.0% | **97.4%** | 0.2% | 0.0% | 0.7% | 0.0% | 2.0% |
| | Clear water | 0.0% | 0.5% | **99.8%** | 0.0% | 0.0% | 0.0% | 0.0% |
| | Mine+Tailings | 0.0% | 0.7% | 0.0% | **100%** | 0.0% | 0.0% | 0.0% |
| | Agricultural | 0.0% | 0.2% | 0.0% | 0.0% | **98.5%** | 0.3% | 0.8% |
| | Built-up | 0.0% | 0.2% | 0.0% | 0.0% | 0.5% | **99.7%** | 0.0% |
| | Grassland | 0.0% | 1.0% | 0.0% | 0.0% | 0.3% | 0.0% | **97.1%** |

**Table 3. Confusion matrix after classification of "after" image**

| | | Actual Value | | | | | | |
|---|---|---|---|---|---|---|---|---|
| | | River | Forest | Clear water | Mine + Tailings | Agricultural | Built-up | Grassland |
| **Predicted** | River | **98.7%** | 0.0% | 0.0% | 0.2% | 0.0% | 0.0% | 0.0% |
| | Forest | 0.7% | **96.3%** | 0.5% | 0.8% | 0.7% | 0.2% | 2.8% |
| | Clear water | 0.0% | 0.3% | **99.5%** | 0.0% | 0.0% | 0.0% | 0.0% |
| | Mine+Tailings | 0.3% | 0.3% | 0.0% | **98.4%** | 0.0% | 0.0% | 0.0% |
| | Agricultural | 0.0% | 0.2% | 0.0% | 0.3% | **97.5%** | 0.0% | 0.2% |
| | Built-up | 0.0% | 0.2% | 0.0% | 0.0% | 0.3% | **99.8%** | 0.0% |
| | Grassland | 0.3% | 2.6% | 0.0% | 0.3% | 1.5% | 0.0% | **97.0%** |

The most important effect is the damage caused by the expansion of the mine & tailings area, which denotes contamination, and this specific classification is highly accurate so can be used to identify which areas were inundated and thus provide an accurate estimate of each type of land affected.

The map that represents this effect is on Figure 7.

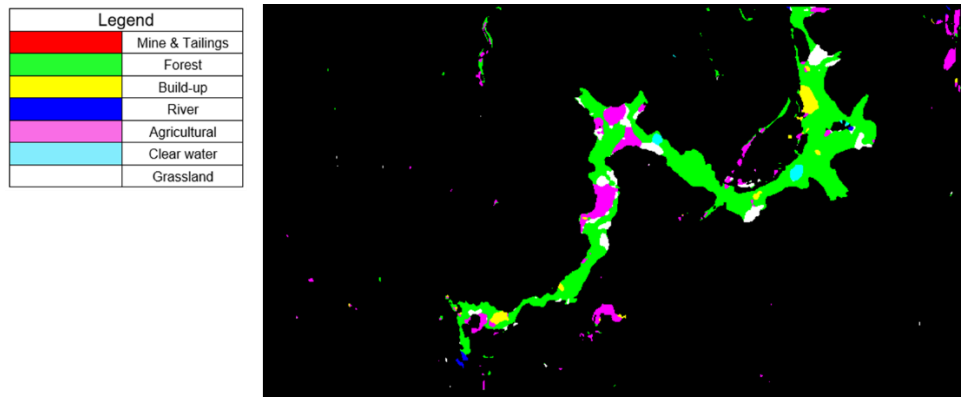

Figure 7. Change map, showing land classes affected

An analysis of the change map indicates that the affected area covered 250 hectares (1 ha = 10,000 m$^2$ = 2.47 acre) with the following impact (Table 4).

**Table 4. Estimation of impact to region by tailings**

|  | hectares |
|---|---:|
| **River** | 1.4 |
| **Forest** | 177.4 |
| **Clear water** | 3.6 |
| **Mine+Tailings** | 0.0 |
| **Agricultural** | 39.0 |
| **Built-up** | 9.8 |
| **Grassland** | 23.0 |
| **Total** | **254.25** |

As shown, the application of the methodology allows for an accurate evaluation of the areas affected by the dam collapse. In practice, knowing the location, type and approximate area of effect is highly valuable, and continuous monitoring can provide time-lapse snapshots of both the expansion of the collapse and also the effect of any remediation measures taken.

## CONCLUSIONS

In this paper, a new approach to use CNNs in order to determine impacted areas after a dam collapse with high accuracy was introduced. Given the characteristics of the problem, the Jigsaw algorithm was used to achieve high accuracy in recognizing mine and tailings regions, to produce change maps that can be applied to impact assessment, response or remediation.

The methodology uses a streamlined approach to quickly acquire, annotate, and make available for training high quality satellite data from ESO's Sentinel-2 mission. It also applies the Jigsaw network to effectively exploit an architecture that makes good use of the surrounding areas of the annotated pixels and extracts efficiently the information of the 12 multispectral bands of the Sentinel-2 data, converging to accuracies of 99% and 98% for seven land use classes. Finally, this study demonstrates a way to use the predictions from the network to build a change map of the affected areas that shows the types and extension of regions affected by a tailings dam collapse.

Future research can use the network and approach to estimate changes in time, thus helping to assess the expansion or effectiveness of remediation of the affected areas. An enhancement would be to define more relevant classes (for example, rural dwellings or roads).

Additionally, using more historical data, the network should increase its robustness due to better generalization across seasons and other image related variations.

Finally, the data can be augmented with ratios between relevant layers in the multispectral bands, to better indicate Vegetation (NDVI) (Kriegler et al., 1969), Water (NDWI) (McFeeters, 1996), Mud (NDMI) (Bernstein, Jin, Gregor, & Adler-Golden, 2012) and other biophysical features.

## APPLICATION OF THE TECHNOLOGY

Dam monitoring and collapse assessment are two ideal applications of this technology. They take advantage of the high accuracy of the classification of tailings/mud and how they affect the previous land use/land cover classification.

Land use / land cover has many other applications in mining, for example, to monitor environmental, economic, and ecological impacts. As such, there can be a business case to maintain a LULC map updated at all times around mining properties. This basic classification can be the basis for any human labeled "before" image for this paper's approach. The system itself runs in a desktop computer with an NVIDIA GPU in a few hours, so this can also be implemented in a cost-effective way in mining companies.

From a practical standpoint, the approach has some limitation due to the source of the data, and the need for human intervention to achieve higher accuracies.

The source of data chosen, ESO Sentinel satellites, have the advantage of being freely available, but for a commercial application are limited because miners require, at the very least, 24-hour response to an event of this category. Nevertheless, the satellite data can be complemented with data capture by private companies that do aerial surveys by plane and even with drones. One alternative would be to use agricultural multi-spectral cameras mounted on drones, which would better assess the impact on forest, agricultural and even water bodies quickly and at a reasonable cost.

For a more accurate "after" labeling, human intervention is also required and will take time but, as described above, we only need around 600 pixels labeled for each class. For an updated "tailings" class, this is equivalent to labeling at most 6 hectares of new data (600 pixels, each 10x10 $m^2$).

Finally, the same approach could be used to monitor and communicate the results of any remediation of the impact in time, allowing for fact-based discussions within the company and with other stakeholders.